\pdfoutput=1

\documentclass[a4paper, 10pt, conference]{ieeeconf}      
\usepackage{FG2017}

\FGfinalcopy 

\IEEEoverridecommandlockouts                              
\usepackage{graphics} 
\usepackage{subfig}
\usepackage{epsfig} 
\usepackage{mathptmx} 
\usepackage{times} 
\usepackage{amsmath} 
\usepackage{amssymb}  

\title{\LARGE \bf
Fast, Dense Feature SDM on an iPhone
}

\author{\parbox{16cm}{\centering
    {\large Ashton Fagg$^1$$^2$, Simon Lucey$^2$$^1$, Sridha Sridharan$^1$}\\
    {\normalsize
    $^1$ Queensland University of Technology, Brisbane, Queensland, Australia\\
    $^2$ Carnegie Mellon University, Pittsburgh, PA, USA}}
    \thanks{This work was supported by QUT and CMU whilst Ashton Fagg was a visiting scholar at CMU.}
}

\begin{document}

\ifFGfinal
\thispagestyle{empty}
\pagestyle{empty}
\else
\author{Anonymous FG 2017 submission\\-- DO NOT DISTRIBUTE --\\}
\pagestyle{plain}
\fi
\maketitle

\begin{abstract}
In this paper, we present our method for enabling dense SDM to run at over 90 FPS on a mobile device. Our contributions are two-fold. Drawing inspiration from the FFT, we propose a Sparse Compositional Regression (SCR) framework, which enables a significant speed up over classical dense regressors. Second, we propose a binary approximation to SIFT features. Binary Approximated SIFT (BASIFT) features, which are a computationally efficient approximation to SIFT, a commonly used feature with SDM. We demonstrate the performance of our algorithm on an iPhone 7, and show that we achieve similar accuracy to SDM.
\end{abstract}

\section{INTRODUCTION}


Xiong \& De la Torre \cite{SDM} proposed an approach to
object \& image alignment known as the Supervised Descent Method
(SDM). SDM shares many similar properties to the Inverse Compositional Lucas Kanade algorithm \cite{lk}
\cite{baker2004lucas}, as it attempts to model the relationship
between appearance and geometric displacement using a sequence of
linear models. SDM \cite{SDM} and its computationally efficient derivatives \cite{ren2014face} \cite{chera} \cite{tzimiropoulos2015project} are considered to be among the state of the art for facial landmark alignment. 

However, SDM relies upon the application of expensive
feature extractions and a large regression matrix. This poses
two issues. First, feature extraction is computationally expensive, which for implementation on mobile devices
may present an unapproachable barrier. Second, there is a cost associated with applying the regression transform. This cost is quadratic relative to the number of landmarks. On mobile devices, where memory bandwidth is limited, this presents a practical barrier for real time performance. 

Recently, there has been interest in simplifying the feature
extraction process by way of learning the regressors on simple
binary features. Binary features are a paradigm which have been explored in depth in other areas, and there are many well known variants \cite{lbp} \cite{brisk} \cite{brief}.

What makes binary features efficient, is that they are derived from localized intensity comparisons, rather than through expensive operations (such as gradients and binning). \cite{ren2014face} proposes an adaptation of previous binary feature representations, based upon random forests, and demonstrates than an SDM-like regression framework can be learned that is more approachable to mobile devices. By producing sparse-binary features, the application of the regressor matrix can be computed efficiently by vector addition, rather than a costly matrix multiplication. \cite{ren2014face}

 The main difference between the assumption of \cite{ren2014face} and \cite{SDM} is the number of pixels are modelled. SDM assumes that each pixel is connected with every other pixel. Whereas, \cite{ren2014face} trades modelling all of these interactions for speed, by only modelling a subset of local interactions.

Whilst the case for \cite{ren2014face} can be made on speed alone, we believe that a more ``middle of the road'' trade-off can be made. Hitherto, SDM, is viewed as a ``one size fits all'' approach. That is, the objective is not
tailored to suit the target device with respect to vectorization or SIMD hierarchies. We believe that the insights from \cite{ren2014face} can be used to break this assumption, and result in a system which is faster than SDM without trading off accuracy as in \cite{ren2014face}.

The key insight from \cite{ren2014face} are that cheap features and careful objective structure can allow for fast execution on mobile devices. The latter is a problem which has already been explored greatly in past literature. Specifically, we draw our inspiration from the Fast Fourier Transform (FFT) \cite{cooley1965algorithm}, which utilizes a property known as Sparse Composition to simplify the execution of the Discrete Fourier Transform in a computationally efficient manner.

Secondly, drawing upon the feature transform of \cite{ren2014face}, we present an approximation to SIFT, called \emph{Binary Approximated SIFT} (BASIFT), which enables similar alignment accuracy to regular SIFT at a reduced computational expense.

Our main contributions are:

\begin{itemize}
\item{We propose Sparse Compositional Regression for SDM and demonstrate its effectiveness in terms of computational cost and alignment accuracy.}
\item{Following the lead of \cite{ren2014face}, we propose a cheap, dense feature which we call \emph{Binary Approximated SIFT} (BASIFT), which is a fast approximation to regular SIFT.}
\item{We show that combining our Sparse Compositional Regressors and BASIFT features results in a significant speed up with little loss in alignment performance compared to standard SDM whilst remaining tractable on mobile devices.}
\end{itemize}

\section{PRIOR ART \& MOTIVATION}



Previously, cameras on mobile devices have largely been restricted to 30 frames/sec. Recently, it has become common to find smartphones and tablets equipped with cameras capable of 60, 120 or 240 frames/sec. In order to leverage the advantages that higher frame rate capture can offer, the requirement for real time process has, at a minimum, doubled.

On desktop hardware, this perhaps is not such a big issue. However, for mobile devices this presents a fundamental challenge. Our argument, is that the alignment objective can be formed in a way that is more applicable to execution on mobile devices with little loss in alignment performance.

\subsection{Supervised Descent Methods (SDM) for Face Alignment}

Supervised Descent Methods \cite{SDM} \cite{saragih2006iterative} draw
strongly from non-linear optimization methods, which have been
classically applied to the face alignment problem. Supervised Descent Methods are
closely related to Gradient Descent Methods, in that they
employ a linearized estimate of the parameters and refine this
estimate iteratively until some convergence criteria is reached.

More particularly, SDM learns a set of generic descent directions from real data, which
can be used to form a linear relationship between the geometric
displacement of a set of points and the appearance within the image. \cite{SDM}

The beauty of this approach is that it does not require the inversion
of a large, generative model in order to form the estimate. \cite{cootes2001active} \cite{matthews2004active}

To the best of our knowledge, SDM (and similar methods) are widely considered state of the art in
2D facial landmark alignment. The SDM objective is typically posed as:
\begin{equation}
  \mathbf{R}^{(k)} = \arg \min_{R} \sum_{n=1}^{N} \sum_{\Delta \mathbf{x} \in \mathbb{X}^{(k)}} \|\Delta \mathbf{x} - \mathbf{R}\Phi(\mathbf{x}_n^{*} + \Delta \mathbf{x})\|
\end{equation}

where \(\mathbb{X}^{(k)}\) is the set of perturbations from
ground-truth landmarks \(\mathbf{x}_n^{*}\) after applying the SDM to
the previous \(k-1\) regressors, \(N\) is the number of examples in
the training set.

For applying SDM on low power devices and for large numbers of points,
there are two distinct drawbacks which must be addressed.

\subsubsection{Feature Cost} SDM requires that features be extracted
over local patches at all points around the face for every stage of
the fitting process (i.e. prior to application of each
regressor). Traditionally, SIFT  \cite{sift,SDM} features have been favored in
past work. Whilst SIFT features are effective and allow the model to
generalize well, they are expensive to compute, despite approximations frequently being made in practice \cite{vlfeat}.

\subsubsection{Regression Cost} As the number of landmarks
increases, the regression matrix becomes large. Not only
does this require a large amount of memory to store, applying the
matrix transform to the feature vector incurs a significant CPU cost. This
is problematic on mobile devices, where memory, CPU and
energy resources are limited. Mobile devices also tend to have limited CPU cache, which in turn becomes the limiting factor where the regression problem is large as cache misses occur frequently.

\begin{figure}
    \centering
    \includegraphics[scale=0.45]{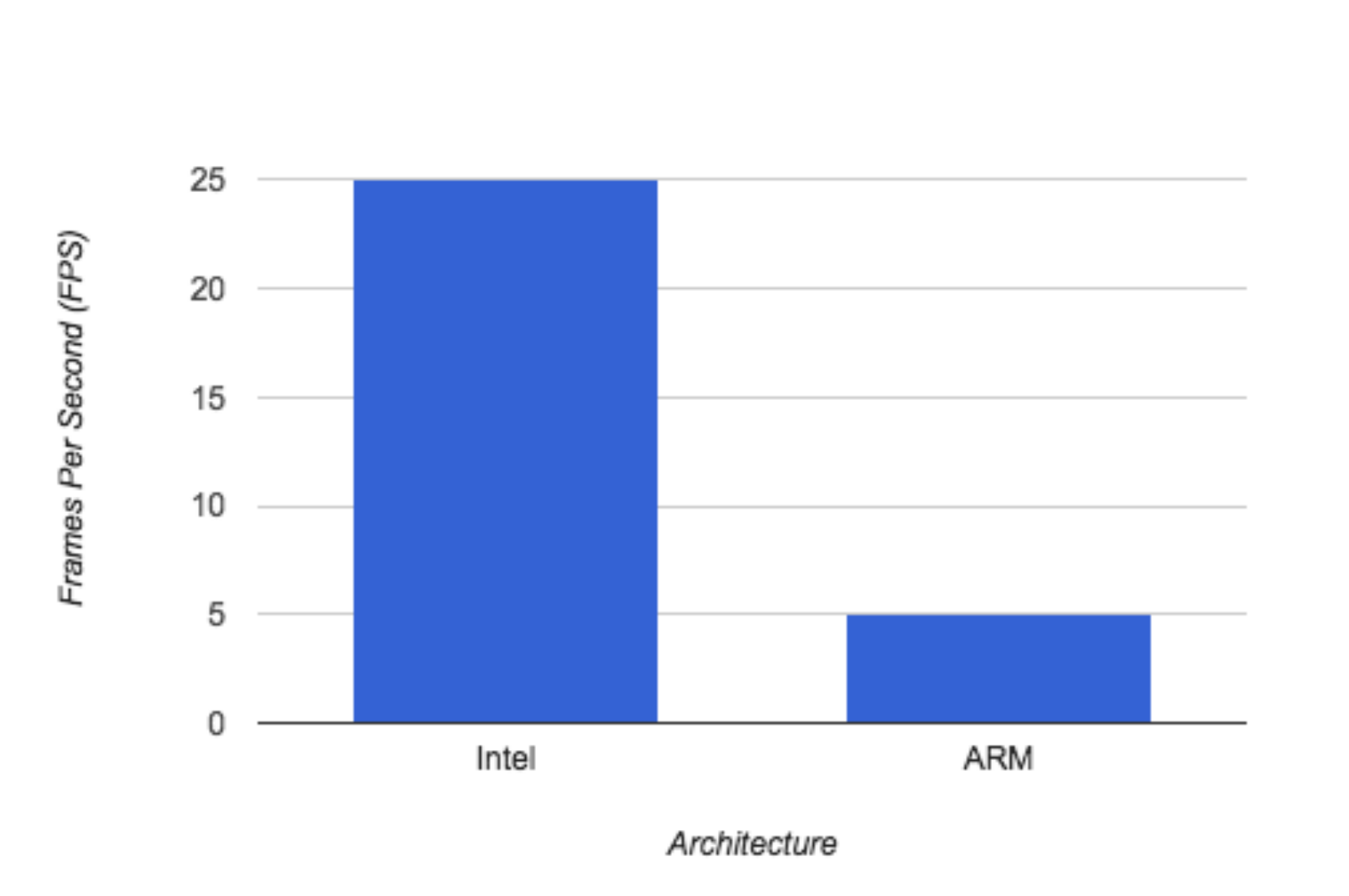}
    \label{fig:speeddensesdm}
    \caption{A comparison of speed of classical SIFT SDM on Intel (Core i7) and ARM (Apple A10) architectures. As one can see, classical SDM is intractable on ARM.}
\end{figure}

\subsection{Binary Features}


In past work, there has been some interest in replacing expensive feature transforms with computationally efficient alternatives. One of the most well explored paradigms in this area is binary features. Binary features are considered computationally efficient, as they performance only basic spatial comparisons (such as \(<\)) rather than expensive multiplications or convolutions. Most modern CPUs are able to perform such comparison operations very efficiently, which makes binary features a compelling option for applications demanding efficiency. Well known examples of binary features are the Local Binary Pattern (LBP) \cite{lbp}, BRISK \cite{brisk} and BRIEF \cite{brief} feature representations. The key insight from binary features is that a large amount of ordinal information can be compactly encoded. Binary features have been shown to work well in a number of applications in Computer Vision.


Recently, \cite{ren2014face} proposed a new form of binary feature specifically designed for facial landmark alignment. Based around random forests, a number of pixel comparisons are performed at randomly selected indices around the landmarks. Since there is a predictable number of states, the binary pattern from a single tree can be compactly encoded as a vector consisting of only a single active element (a 1 placed in the position corresponding to the state). The overall feature vector is formed from the concatenation of all local features over the face. Depending on the number of trees and comparisons selected, the feature vector may be very large. However, because only a small number of elements are marked as active, it will be very sparse.

The sparse nature of the features also lends a computational advantage to the regression cost. Since only a small number of elements are active, the regression transform can be applied as a series of vector additions, rather than a traditional matrix multiplication. Since no multiplication operations are needed, and the vector addition method lends itself well to SIMD hierarchies on modern chipsets, the result is a significant speed up.

In \cite{ren2014face}, it is proposed that a trade-off between speed and accuracy can be made by adjusting the number of trees (in effect changing the number of pixel interactions modelled). Specifically, \cite{ren2014face} demonstrates two variants: a fast variant and a slower, more accurate version. The fast version is reported to run at approximately 3000 FPS. Whereas the slower version, while achieving only 300 FPS, more accurate.

By their nature, predictors formed from random forest based sampling methods are susceptible to noise. As such, \cite{ren2014face} proposes a validation step to select the optimal interactions to model. Since natural images possess a high degree of local correlation, modelling only a few interactions may not capture an optimal amount of ordinal information and such measurements may be highly susceptible to noise.

Core to our argument, is that dense features are able to better generalize, as they capture all local interactions. The core problem, however, is that many dense features are computationally expensive to extract and the feature structure cannot be leveraged to allow for efficient application of the regression transform. It is this problem which we shall explore in this paper.



\subsection{Sparse Composition}



The Fast Fourier Transform (FFT) \cite{cooley1965algorithm} is a very common 

When the FFT is applied to a signal, the signal is transformed from the time domain to the Fourier (frequency) domain.

The reason why the FFT is used is due to its superior computational
performance compared with the Discrete Fourier Transform (DFT). The
FFT can be posed as a matrix transform:

\begin{equation}
  \hat{\mathbf{z}} = \mathbf{F}_D \mathbf{z}
\end{equation}

where \(\mathbf{z}\) is a vectorized, \(D\) dimensional signal and
\(\hat{\mathbf{z}}\) is it's representation in the Fourier domain. \(\mathbf{F}_D\) is a \(D \times D\) discrete one dimensional Fourier transform basis.

Naively, since \(\mathbf{F}_D\) is a dense matrix, the cost of this
operation should be \(O(D^2)\). However, since the seminal work of
Cooley \& Tukey \cite{cooley1965algorithm}, it has been well understand that \(\mathbf{F}_D\) has
intrinsic redundancies which can be leveraged in order to reduce the
computational cost. This property is what we call sparse composition. A
compositionally sparse matrix can be defined as:

\begin{equation}
  \mathbf{R} = \prod_{l=1}^L \mathbf{S}_l
\end{equation}

\(\mathbf{R}\) is a dense matrix, which is composed as the product of
a set of matrices, \(\{\mathbf{S}_l\}^L_{l=1}\) which individually
are sparse or group sparse. At runtime, a speed up can be obtained when one evaluates the product as:

\begin{equation}
  \hat{\mathbf{z}} =  (\mathbf{S}_L(\ldots(\mathbf{S}_2(\mathbf{S}_1 \mathbf{z}))))
\end{equation}

rather than the more expensive:

\begin{equation}
  \hat{\mathbf{z}} = (\mathbf{S}_l \ldots \mathbf{S}_2 \mathbf{S}_1) \mathbf{z}
\end{equation}

The reason for this will become clear when we consider the FFT. Classically, the FFT
complexity is defined as \(O(D \text{log} D)\), rather than the naive
\(O(D^2)\).

\begin{figure*}
  \centering
  \includegraphics[scale=0.75]{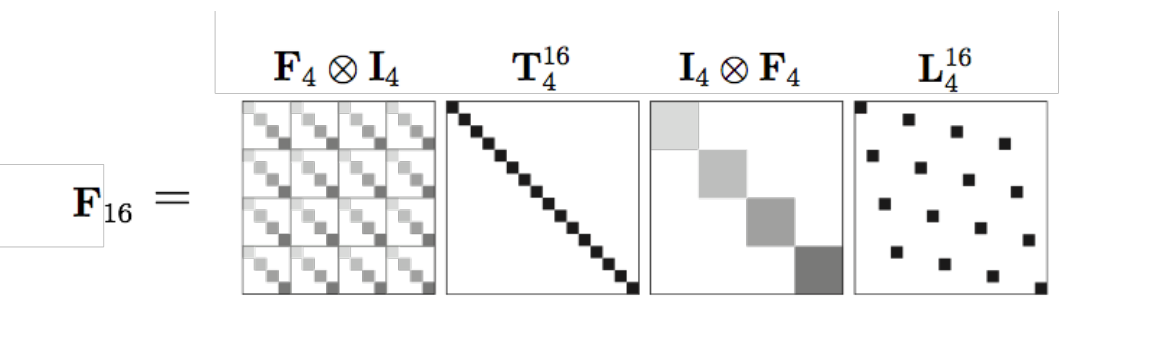}
  \label{fig:fft}
  \caption{The 16 dimensional FFT is formed compositionally as a number of subsequent, small matrix transforms rather than a single, potentially large matrix transform. This is a well known example of a sparse composition, and this structure is intrinsic to obtaining such a significant reduction in computation time. Figure adapted from \cite{franchetti2009discrete}.}
\end{figure*}

Core to this insight, also, is the structure and execution process followed at run time. What many do not realize, is that the FFT is executed differently depending upon the architectural strengths of the particular device. Many popular software libraries, such as the Fastest Fourier Transform in the West (FFTW) \cite{fftw}, perform different computations in order to optimize the speed of the algorithm for different architectures. The sparse compositional form of the FFT, allows for different forms to be formulated \cite{franchetti2009discrete, franchetti2006fft} or even generated automatically \cite{meng2010spiral, puschel2005spiral} to suit different computational architectures.

An obvious question is why the sparse compositional form is faster than a single matrix transform. The answer is two-fold. First, by introducing structure into the matrices, the matrix multiplication can be efficient evaluated. For example, multiplication of block-sparse matrices is able to be executed in parallel and results in less operations compared to a naive multiplication. Second, by sizing the components appropriately and reusing the intermediate products, we enable the top-level CPU caches to be used more efficiently. As such, the number of cache misses incurred is reduced.

It is this insight from which we primarily draw our inspiration: sparse composition can enable efficient application of linear transforms in a device-centric manner. For facial landmark alignment, we believe a similar compositional structure can be enforced to achieve a significant improvement in fitting-time performance with minimal loss in alignment accuracy.

\section{OUR METHOD}
\subsection{Objective \& Structure}

The general form of our objective is defined as:

\begin{equation*}
    \arg\min_{\mathbf{S}_1, \ldots, \mathbf{S}_L} \sum_{n=1}^N \sum_{\Delta \mathbf{x} \in \mathbb{X}^{(k)}}\|\Delta \mathbf{x} - \prod_{l=1}^L \mathbf{S}_l \Phi(\mathbf{x}_n^{*} + \Delta \mathbf{x})\|^2_2 \\
\end{equation*}

where \(\mathbf{S}_l \in \mathbb{M}_l \forall l = 1, \dots, L\).

Rather than a single regression matrix, we form our regression transform compositionally, with each \(\mathbf{S}_l\) being a component in this composition. We specify that each component can have individual structure and their respective sizes be selected such that cache misses are minimized. This implies that each component is a member of a set of structured matrices, \(\mathbb{M}_l\). This structure could be, but not limited to, block sparse \cite{eldar2010block}, diagonal, Toeplitz or circulant \cite{hariharan2012discriminative,valmadre2014learning} as these structures provide redundancies which enable multiplication to be performed efficiently.

The key to selecting the component sizes and structures depends largely upon the target architecture. However, we advocate that the first component should be the largest and most sparse, with each subsequent transform being smaller and less sparse. The reasoning for this is that if the first component is able to reside within the top level memory cache, the memory bandwidth remains constant for the rest of the expression. Ideally, the result of each subsequent transform should be able to reside entirely in the top-level cache. As the dimensionality of the intermediate result is shrinking, the cache requirement decreases with each subsequent component. We shall explain our choice of structure in the following section.

For our experiments, we adopt a three component form, which we formulate based on our selected 49 point facial model. In traditional SDM, the single dense regressor implies connectivity at every point within the regression model with every other point. In other words, points lie at opposite ends of the face are assumed to influence one another. Conversely, \cite{ren2014face} assumes that adjacent points have no influence upon one another. 

It is our thesis that in order to achieve good performance (both accurate and fast), we need to break both of these assumptions and adopt an approach which focuses upon all of these aspects at once. Instead, we propose a three component form which can describes local, neighbourhood and global connectivity as individual components.

Locally, we assume that each patch has certain interactions which pertain only to itself. In the neighbourhood component, we group related points together and allow interactions to form between adjacent points. At the final level, we enforce a global connectivity to boost allow all neighbourhoods to interact with one another. By making such assumptions, we can form our objective in such a way that it can be executed efficiently in a similar vein to the FFT.

We define our objective as:

\begin{equation*}
    \arg\min_{\mathbf{S}_1, \mathbf{S}_2, \mathbf{S}_3} \sum_{n=1}^N \sum_{\Delta \mathbf{x} \in \mathbb{X}^{(k)}}\|\Delta \mathbf{x} - \prod_{l=1}^3 \mathbf{S}_l \Phi(\mathbf{x}_n^{*} + \Delta \mathbf{x})\|^2_2 \\
    $$
    \text{        subject to: }\\
    $$
    \mathbf{S}_1 \in \mathbb{M}_1^{D_1 \times D_F}, \mathbf{S}_2 \in \mathbb{M}_2^{D_2 \times D_1}, \mathbf{S}_3 \in \mathbb{M}_3^{2P \times D_2} \\
\end{equation*}

\begin{figure*}[t]
    \centering
    \subfloat[Component 1 Structure]{\includegraphics[scale=0.20]{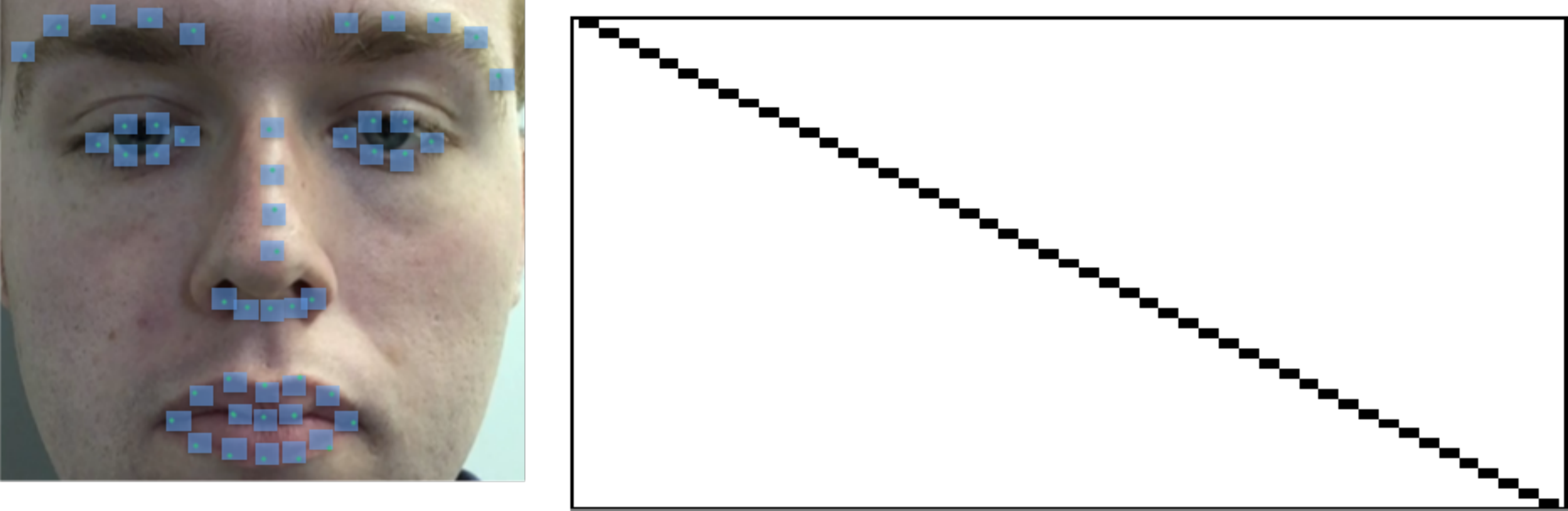}}\hspace{25pt}
    \subfloat[Component 2 Structure]{\includegraphics[scale=0.20]{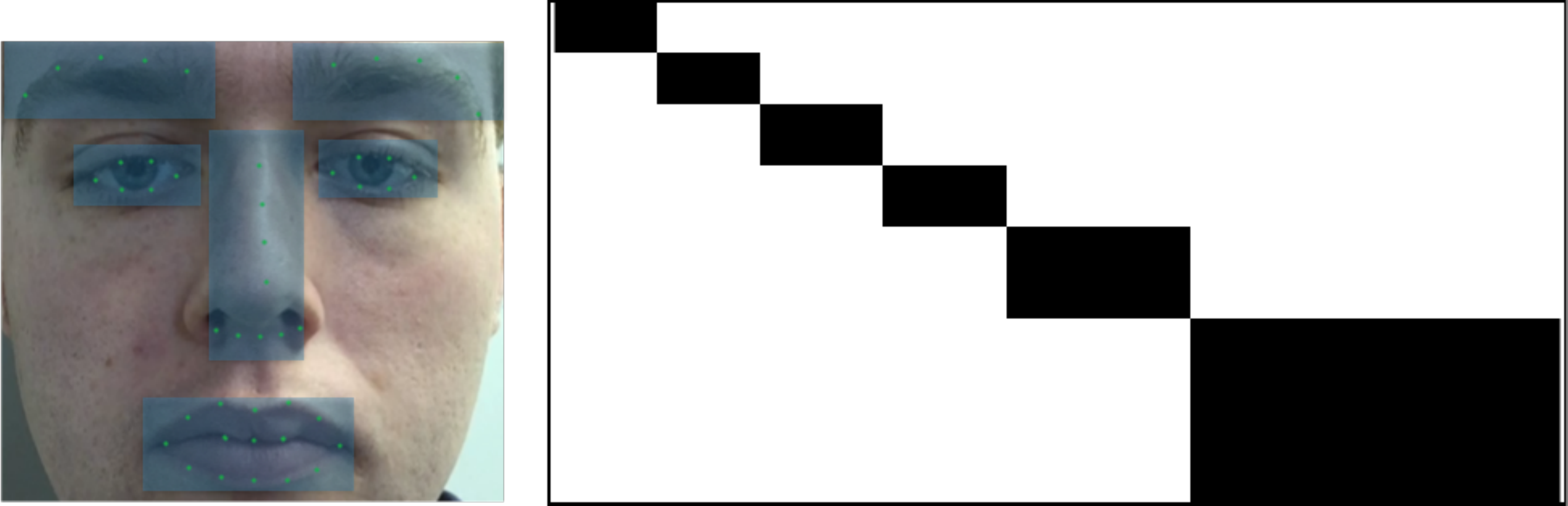}}\\
    \subfloat[Component 3 Structure]{\includegraphics[scale=0.20]{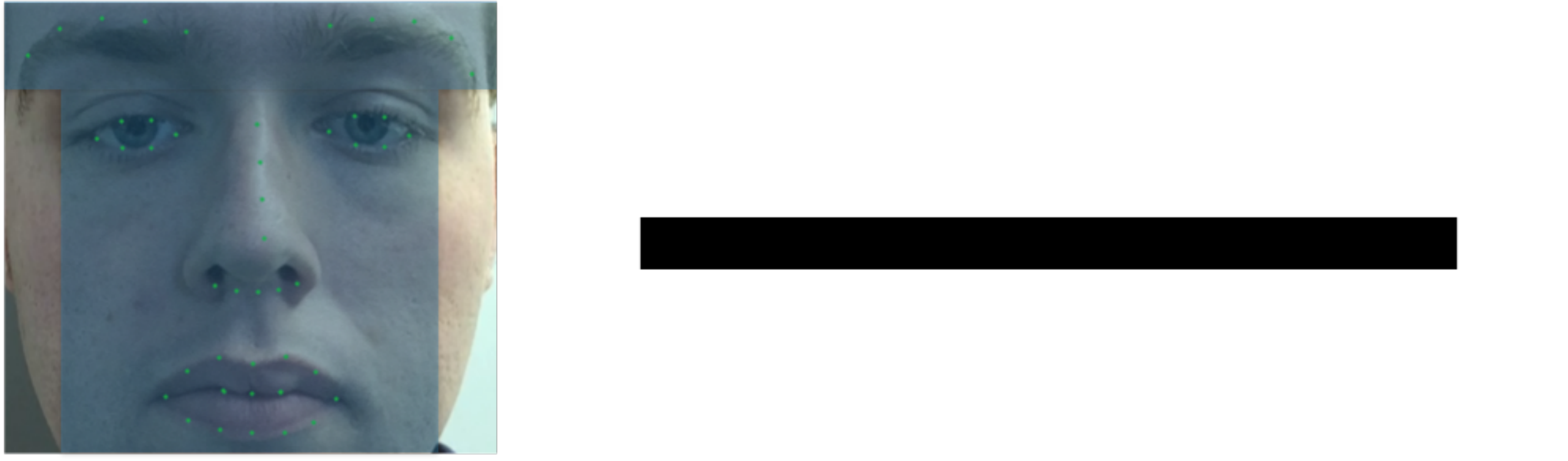}}\hspace{25pt}
    \subfloat[Composition]{\includegraphics[scale=0.28]{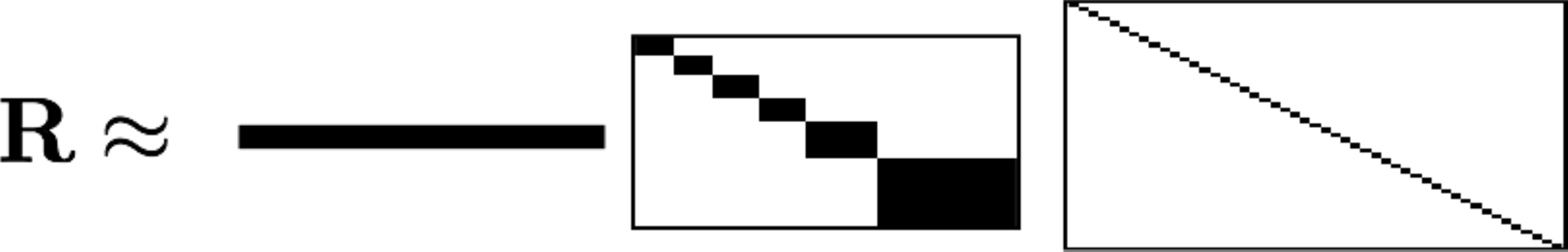}}
    \caption{Sparse Structures for Components 1 and 2. Component 1 (a) groups local features pertaining to only their host patch. Whereas, Component 2 (b) groups features over local neighbourhoods. Component 3 (c) is fully connected, and serves as a rectification layer. (d) Demonstrates how our method works in practice, we essentially replace the standard SDM regression matrix with the product of our 3 components.}
    \label{fig:comp1comp2}
\end{figure*}

\subsubsection{Component 1}

The structure of our first component enforces strict local connectivity around the facial landmarks. At each landmark, we impose a local feature dimensionality reduction transform, which we learn at training time for each individual point (meaning that we assume that each landmark cannot be reduced by the same transform). When using SIFT features, the features corresponding to each landmark is reduced to a 16 dimensional vector from a 128 dimensional vector. For 49 points, the total feature vector length is 6272 (forming the number of total columns in \(\mathbf{S}_1\)), post-reduction this length is 784 (forming the number of total rows in \(\mathbf{S}_1\)).

This structure is illustrated in Figure ~\ref{fig:comp1comp2}(a). In total, only approximately 2\% of the elements are non-zero. This is the most sparse of all components.

\subsubsection{Component 2}

The structure of the second component relaxes the local connectivity assumption to draw connections between local point neighbourhoods. That is, we emphasise that groups of points can strongly influence one another (all mouth points, left eye, right eye, nose and eyebrows), but there is no cross-neighbourhood interaction. We illustrate this structure in Figure ~\ref{fig:comp1comp2}(b), which results in approximately 20\% of the elements being non-zero.

In terms of sizing, we reduce the 16 dimensional local features down to 8. This is a 50\% reduction from the previous component.

\subsubsection{Component 3}

For the final component, we enforce no structure, meaning that can be assumed to be completely dense. This is effectively a rectification step to map the output to the final point update estimate. This is the smallest matrix in terms of dimensionality, but the least sparse. This is important as unlike the previous components, we cannot exploit any structure in order to efficiently evaluate the matrix transform.

\subsection{Choosing component structure}

An obvious question is how one would select an appropriate structure. The main concern is introducing structure which allows modern CPUs to cleverly perform the matrix multiplications.

In the case of components 1 and 2, we adopt a block sparse structure. The key difference being is that the block size in component 2 is variable, whereas in component 1 it is of a fixed size. 

If we examine Component 1, what we notice is that although it is large in dimensionality, it is highly sparse. Also, the interactions are purely local. This lends itself well to modern computing architectures, since only a small number of interactions need to be computed (since the matrix is sparse), and those that do need to be computed can be computed independently (taking advantage of SIMD instructions and multiple cores). Once will also note, that the size of the individual blocks along the diagonal are sized to fit neatly within SIMD registers. Since the block sizes are multiples of powers of 2, it can easily be divided into 4, 8, 16 or 32 computations at run time. 

Component 2 shares this same structure, with different grouping rules. Instead of grouping only elements related to single patches, we extend our local neighbourhoods to include information from points which are likely to directly interact. For example, we group all mouth points together. Despite the grouping being of varying sizes, this structure is still able to be exploited at run time to enable efficient multiplication to be performed. Even with the larger groups, it can be performed in parallel and will still be able to broken into multiple steps to neatly fit SIMD registers.

If, for example, we were to arbitrarily define what the local interactions should be, there is no guarantee that it will be computationally efficient (since interactions may no longer be adjacent). This adjacency also results in a decrease in memory bandwidth, since taking into account sparsity, blocks can be loaded quickly due to not needing to make large strides across the address and register space (assuming attention is paid to storage order of the components).

We have formulated our objective such that it works well on both Intel and ARM architectures. We are not claiming that this form is optimal in either case. What we wish to demonstrate by using this form is that simple sparse composition can provide a performance increase with minimal loss in alignment performance.

\subsection{Training \& Validation}

Since SDM is traditionally solved as a Ridge Regression objective, an
L2 regularization penalty is frequently used to ensure a suitable solution can be obtained. However, with our SCR objective we are
unable to make this assumption. Due to the multicomponent form, it is
not clear what a suitable strategy would be for applying a Tikhonov regularizer \cite{ng2004feature} to our objective.

Instead, we chose a Gradient Descent method combined with an
early-stopping regularization scheme. In order to solve our objective,
we generate both a training and validation set using a Monte-Carlo
sampling method similar to that of \cite{SDM}. At each iteration, we
update the parameters using gradients derived from the training
set. We monitor the suitability of our solution using the validation
set, and instead of applying a structural regularizer we select the
solution which best minimizes the loss against the validation set.

\subsection{Feature Implementation}

In recent work, there has been interest in replacing expensive feature
transforms with more efficient alternatives. \cite{ren2014face}
Specifically, the features proposed by \cite{ren2014face} while cheap to compute,
are still high dimensional. Even for a modest sized problem, the
resulting feature vector can contain tens of thousands of elements. If
the regression was to be applied naively, this would obviously pose an
issue for mobile devices.

We believe that dense, compact features can provide good performance when approximated efficiently. Although SIFT is traditionally expensive to compute, the feature vectors produced are not overly large (certainly less than that of \cite{ren2014face}), and are more ``cache friendly'' at fit time, particularly for large numbers of points. In the next section, we propose an efficient approximation to SIFT and show experimentally that regressors learned on these new features provide approximately the same registration performance as that of regular SIFT at a considerable reduction in computation. This avenue was investigated, as SIFT seems to provide a good balance between feature dimensionality and registration performance. As such, we deemed that an approximate SIFT which is more computationally efficient would complement our reduced regression cost.

We draw upon the work of \cite{ren2014face, bitplanes} which demonstrates that useful features can be derived by performing simple intensity comparisons. Our proposed approach is to leverage the computational efficiency of these features, whilst approximating the expressiveness that SIFT brings to the alignment problem.

SIFT can be broken down into a number of steps. Broadly speaking, they are:

\begin{enumerate}
    \item{Derive image gradients in x and y}
    \item{Compute gradient magnitudes and orientations}
    \item{Bin gradients according to their orientations}
    \item{A meanshift-like step to encode spatial invariance}
\end{enumerate}

Key to our insight, is that a binning step can be approximated according to a number of bit comparison. If we select 8 orientation bins, 3 comparisons can be used to enumerate which bin a particular position falls in (since \(2^3=8\)). Rather than calling upon expensive trigonometric functions (such as \texttt{atan2}), we propose a simple alternative.

Given gradients \(\mathbf{G}_\mathbf{x}\) and \(\mathbf{G}_\mathbf{y}\), the corresponding orientation bin can be derived by evaluating the following at every spatial position (\(x,y\)):

\begin{equation}
\eta_{x,y} = [\mathbf{G}_\mathbf{x}(x,y) > 0, \text{ }\mathbf{G}_\mathbf{y}(x,y) > 0, \text{ } |\mathbf{G}_\mathbf{x}(x,y)| > |\mathbf{G}_\mathbf{y}(x,y)|]
\end{equation}

The result is a 3-bit binary representation denoting the bin number with which the orientation is associated. Note that our binning is not equivalent to that of normal SIFT. As such, a look-up table which maps our binning back to the standard SIFT binning is required. The reason for this is due to our next step, as ordering is important.

In regular SIFT, the binning step stores both the orientation (which bin) and the magnitude. Instead, we only place a single active element in the orientation bin to denote that a particular spatial position belongs to that bin. The result is a highly sparse, binary vector (similar to \cite{ren2014face}).

Our final step was to approximate the meanshift step of SIFT with a linear regression:

\begin{equation}
    \arg\min_{\mathbf{L}} \|\Phi(\mathbf{x}) - \mathbf{L}\eta(\mathbf{x})\|_2^2
\end{equation}

where \(\mathbf{x}\) is an image, \(\Phi\) is the traditional SIFT operator and \(\eta\) is the binary comparison operator we proposed above. We learned the mapping matrix \(\mathbf{L}\) using random patches generated from the Imagenet dataset. \cite{deng2009imagenet} Our reason behind using ImageNet rather than face patches is that SIFT is not itself face specific.

In a naive approach, \(\mathbf{L}\) would likely need to be stored as a matrix of floating point numbers. To obtain a greater speed up, we store only the sign of \(\mathbf{L}\), which can be stored as a matrix of signed integers. Since the problem now consists entirely of fixed-point elements, the application of the feature transform can be much faster than if floating point numbers were used. We also employ a similar trick to \cite{ren2014face}, and perform this multiplication as an addition rather than a traditional matrix multiplication since \(\eta(\mathbf{x})\) is sparse.


In this form, our feature transform is simply:

\begin{equation}
 \Phi(\mathbf{x}) \approx \text{sign}(\mathbf{L})\eta(\mathbf{x})
\end{equation}

We call our representation \emph{Binarized Approximated SIFT} (BASIFT). In the next section, we present our results based upon BASIFT, which will demonstrate that performance is close to that of using regular SIFT.

\section{EXPERIMENTS}

We compared our method against standard SDM and the approach of \cite{ren2014face} on the 300W dataset \cite{300w}. It is not our goal to achieve state of the art performance, we merely wish to demonstrate that sparse compositional regression can achieve similar levels of performance to current methods.

Since the authors do not make source code for \cite{ren2014face} available, we are using our own implementation. For SDM, we use a modified version of our SCR code, and train on the same data for all of the methods for fair comparison.


We use three variants of \cite{ren2014face}. Variants 2 and 3 are configured in accordance with the settings \cite{ren2014face}, with Variant 2 aiming for accuracy and Variant 3 aiming for speed. Variant 1 was selected to approximately match the runtime performance of our SCR+BASIFT method for fair comparison.

For training, we generate 5 synthetic perturbations from the training set of 300W. For validation, we use 2 synthetic perturbations taken from a random selection of frames from the 300VW dataset. \cite{300vw}

Our reasoning behind using 300VW for validation, rather than simply taking more perturbations from 300W, was to ensure that our models do not over-fit to the appearance of subjects in the training set. Additionally, since we are not using our validation set to select a parameter, it is not possible to partition 300W into a training set and a validation set. In order to maximize the data available for training, and provide a validation set which best emulates real world data, we opted to use two datasets. Whilst this approach is not conventional, in our case we believe it is a reasonable approach to selecting a suitable solution given the constraints we encounter.

For SDM and \cite{ren2014face}, we use the validation set to perform a Golden Section Search to obtain the regularization parameter, \(\lambda\). For SCR, we use the validation set with gradient descent to inform our choice of solution as described in the previous section.

\subsection{Runtime Performance}

We implemented our system using the Eigen \cite{eigen} C++ library in order to allow for easy optimization of mathematical expressions, paying careful attention to storage order of matrices. We opted to use the iPhone 7 for our testing as it is the current state of the art in cellular handsets.

\begin{figure}
    \centering
    \includegraphics[scale=.5]{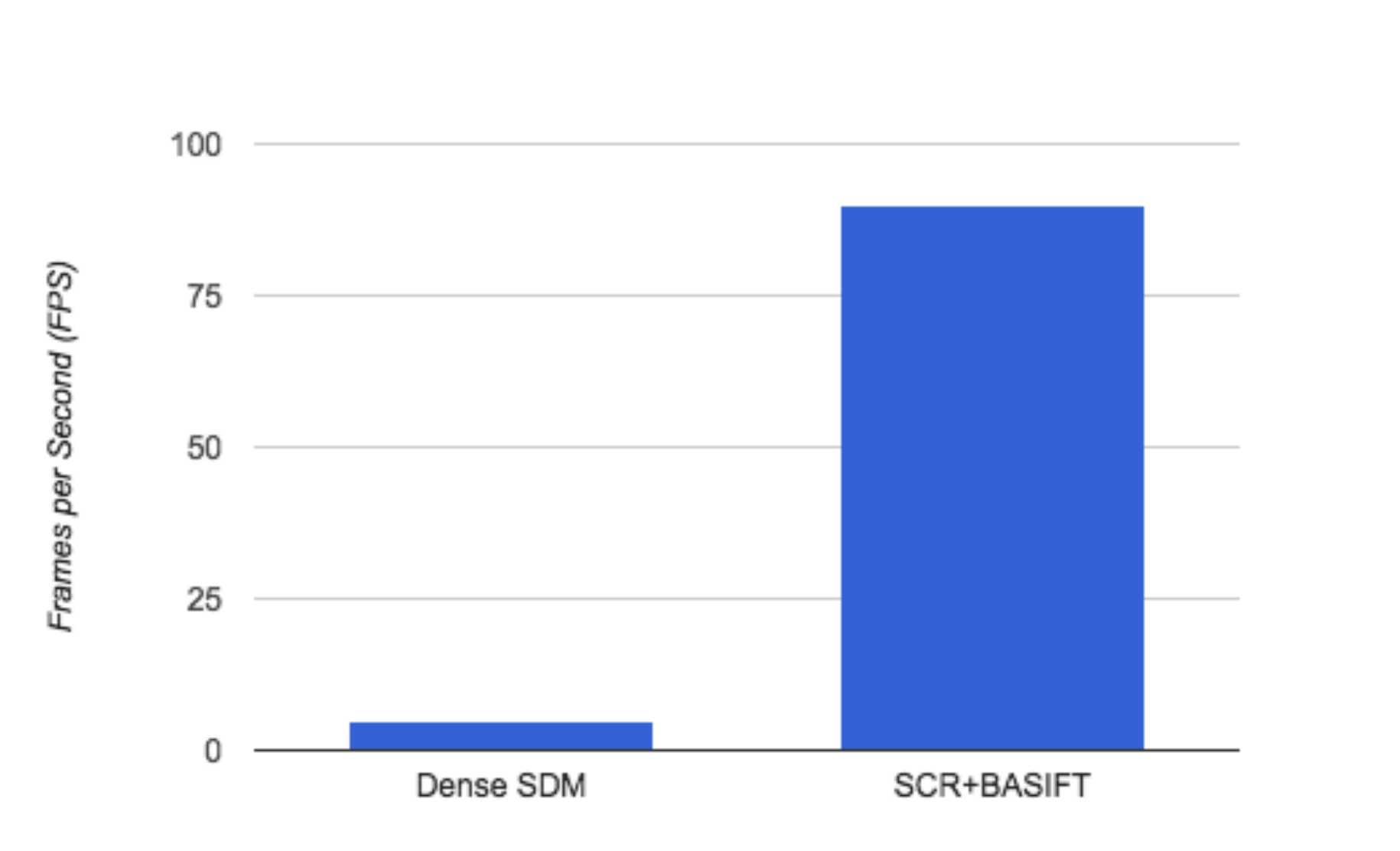}
    \caption{Comparison in execution speed between classical SDM and SCR+BASIFT on the iPhone 7.}
    \label{fig:scrspeedup}
\end{figure}

In this experiment, we combined all components of the system to give an account of our runtime performance on an iPhone 7. We achieved a runtime speed of approximately 90 FPS, for a 5 layer model. Compared to regular dense SDM (which runs on the iPhone at a mere 5 FPS), this is a significant speed up.

\subsection{Fitting Performance}

For a more formal validation of performance, we used the 300W dataset to assess the performance of our method compared with regular SDM and \cite{ren2014face}. For each frame in the test set, we generated 20 random initializations and computed the performance of each method. We keep these random initializations the same for all methods, and use the error metric specified by \cite{300w} (error as a percentage of eye distance).

Our reasoning behind using multiple initializations was to better understand the generalization of the method. It was our hypothesis that the binary features would perform worse than our dense features on average, and have a higher variability in fit quality.

We note that our classical SIFT SDM implementation achieves an average error of 3.22 on 300W.

\begin{figure*}[t]
    \centering
    \includegraphics[scale=.4]{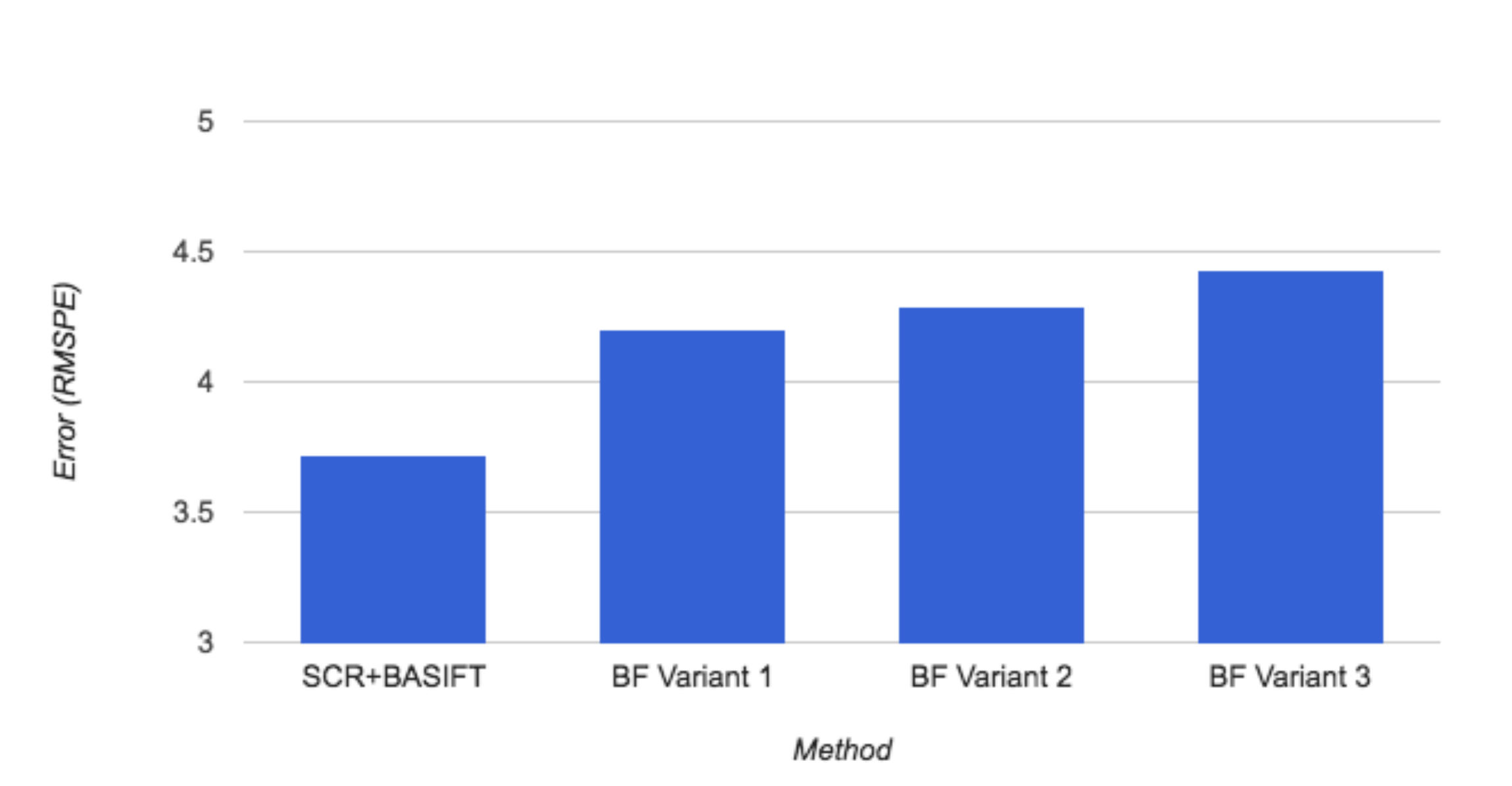}
    \caption{Comparison of methods on 300W. Our method outperforms the binary feature variants of \cite{ren2014face} by a significant amount. Also, given that our classical SIFT SDM implementation achieves an average accuracy of 3.22, we note that SCR+BASIFT is the closest of all the methods to this benchmark. These results are impressive, given that our SCR+BASIFT runs at 90FPS on the iPhone 7, compared to 5 FPS of classical dense SDM.}
    \label{fig:300wperfgraph}
\end{figure*}

In Figure ~\ref{fig:300wperfgraph} we present our results. What is noticeable, is that our method outperforms the binary feature methods. In fact, our method is the closest of all the methods to regular SDM, achieving an average error of 3.72.

What is also clear from these results, is that there is a diminishing return from increasing the number of point interactions modelled by the binary features. 

\begin{figure*}[t]
    \centering
    \includegraphics[scale=.5]{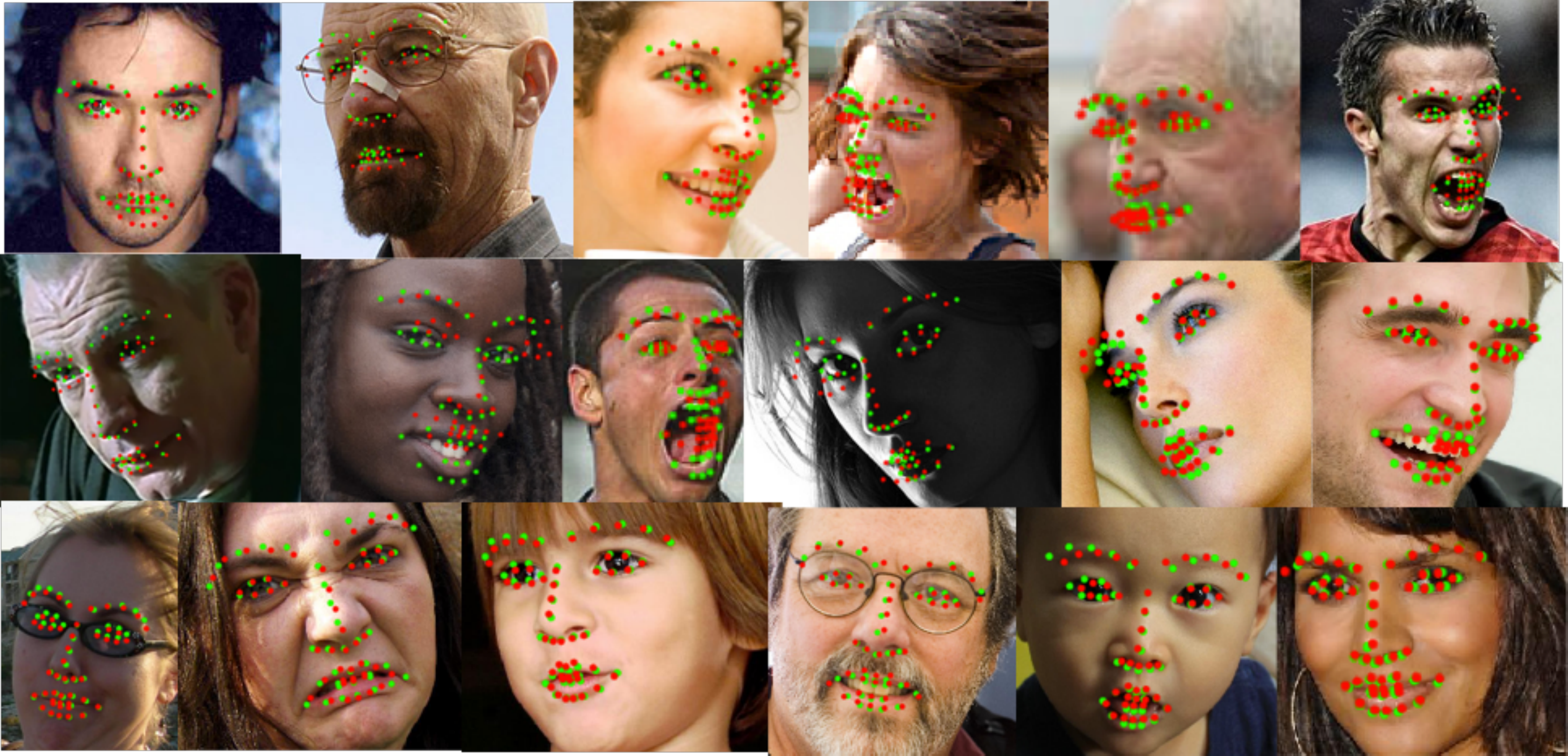}
    \caption{Examples of performance of the test set of 300W. The results of our SCR+BASIFT are in green, whereas the results from \cite{ren2014face} are in red.}
    \label{fig:300wexamples}
\end{figure*}

We include a set of examples from the 300W test set in Figure ~\ref{fig:300wexamples}.

\section{Future Work}

The work in this paper has raised a number of questions which warrant further investigation. 

In this paper we have primarily discussed using Sparse Composition to enable real-time performance of SDM on low power devices. Another such method which could benefit from our insights, is the Global SDM \cite{xiong2015global}, which involves using a set of SDM models to partition the descent space into related regions. For multiview tracking of faces, this involves the application of multiple SDM models. Hence, the application of our insights to the multiview tracking paradigm may result in a speed up.

Another area we have not investigated in this paper is the performance of our proposed method for large numbers of points. Specifically, for dense reconstruction of faces, a larger number of points is required to produce a high fidelity result.

Also, we would advocate further investigation of the effect that different structures have on regression performance and subsequent computational performance on different architectures. Further investigation into how different structures and numbers of components perform on a wider variety of architectures compared to a single regressor would be of interest to the community.

\section{CONCLUSION}

In this paper, we have proposed our method for speeding up SDM on mobile devices: specifically, the iPhone 7. We have shown that our method can achieve similar alignment performance to compared with many state of the art methods, for a significantly reduced computational cost. We also proposed BASIFT, which is a fast, binary approximation to SIFT which lends itself well to mobile devices.

%

\bibliographystyle{ieee}
\bibliography{egbib}

\begin{thebibliography}{10}\itemsep=-1pt

\bibitem{bitplanes}
H.~Alismail, B.~Browning, and S.~Lucey.
\newblock Bit-planes: Dense subpixel alignment of binary descriptors.
\newblock {\em arXiv preprint arXiv:1602.00307}, 2016.

\bibitem{chera}
A.~Asthana, S.~Zafeiriou, S.~Cheng, and M.~Pantic.
\newblock Incremental face alignment in the wild.
\newblock In {\em Proceedings of the IEEE Conference on Computer Vision and
  Pattern Recognition}, pages 1859--1866, 2014.

\bibitem{baker2004lucas}
S.~Baker and I.~Matthews.
\newblock Lucas-kanade 20 years on: A unifying framework.
\newblock {\em International journal of computer vision}, 56(3):221--255, 2004.

\bibitem{brief}
M.~Calonder, V.~Lepetit, M.~Ozuysal, T.~Trzcinski, C.~Strecha, and P.~Fua.
\newblock Brief: Computing a local binary descriptor very fast.
\newblock {\em IEEE Transactions on Pattern Analysis and Machine Intelligence},
  34(7):1281--1298, 2012.

\bibitem{cooley1965algorithm}
J.~W. Cooley and J.~W. Tukey.
\newblock An algorithm for the machine calculation of complex fourier series.
\newblock {\em Mathematics of computation}, 19(90):297--301, 1965.

\bibitem{cootes2001active}
T.~F. Cootes, G.~J. Edwards, and C.~J. Taylor.
\newblock Active appearance models.
\newblock {\em IEEE Transactions on Pattern Analysis \& Machine Intelligence},
  (6):681--685, 2001.

\bibitem{deng2009imagenet}
J.~Deng, W.~Dong, R.~Socher, L.-J. Li, K.~Li, and L.~Fei-Fei.
\newblock Imagenet: A large-scale hierarchical image database.
\newblock In {\em Computer Vision and Pattern Recognition, 2009. CVPR 2009.
  IEEE Conference on}, pages 248--255. IEEE, 2009.

\bibitem{eldar2010block}
Y.~C. Eldar, P.~Kuppinger, and H.~Bolcskei.
\newblock Block-sparse signals: Uncertainty relations and efficient recovery.
\newblock {\em IEEE Transactions on Signal Processing}, 58(6):3042--3054, 2010.

\bibitem{franchetti2009discrete}
F.~Franchetti, M.~Puschel, Y.~Voronenko, S.~Chellappa, and J.~M. Moura.
\newblock Discrete fourier transform on multicore.
\newblock {\em IEEE Signal Processing Magazine}, 26(6):90--102, 2009.

\bibitem{franchetti2006fft}
F.~Franchetti, Y.~Voronenko, and M.~Puschel.
\newblock Fft program generation for shared memory: Smp and multicore.
\newblock In {\em SC 2006 Conference, Proceedings of the ACM/IEEE}, pages
  51--51. IEEE, 2006.

\bibitem{fftw}
M.~Frigo and S.~G. Johnson.
\newblock Fftw: An adaptive software architecture for the fft.
\newblock In {\em Acoustics, Speech and Signal Processing, 1998. Proceedings of
  the 1998 IEEE International Conference on}, volume~3, pages 1381--1384. IEEE,
  1998.

\bibitem{eigen}
G.~Guennebaud and B.~Jacob.
\newblock The eigen c++ template library for linear algebra.
\newblock {\em http:/eigen. tuxfamily. org}, 2010.

\bibitem{hariharan2012discriminative}
B.~Hariharan, J.~Malik, and D.~Ramanan.
\newblock Discriminative decorrelation for clustering and classification.
\newblock In {\em European Conference on Computer Vision}, pages 459--472.
  Springer Berlin Heidelberg, 2012.

\bibitem{brisk}
S.~Leutenegger, M.~Chli, and R.~Y. Siegwart.
\newblock Brisk: Binary robust invariant scalable keypoints.
\newblock In {\em 2011 International conference on computer vision}, pages
  2548--2555. IEEE, 2011.

\bibitem{sift}
D.~G. Lowe.
\newblock Object recognition from local scale-invariant features.
\newblock In {\em Computer vision, 1999. The proceedings of the seventh IEEE
  international conference on}, volume~2, pages 1150--1157. Ieee, 1999.

\bibitem{lk}
B.~D. Lucas, T.~Kanade, et~al.
\newblock An iterative image registration technique with an application to
  stereo vision.
\newblock In {\em IJCAI}, volume~81, pages 674--679, 1981.

\bibitem{matthews2004active}
I.~Matthews and S.~Baker.
\newblock Active appearance models revisited.
\newblock {\em International Journal of Computer Vision}, 60(2):135--164, 2004.

\bibitem{meng2010spiral}
L.~Meng, Y.~Voronenko, J.~R. Johnson, M.~Moreno~Maza, F.~Franchetti, and
  Y.~Xie.
\newblock Spiral-generated modular fft algorithms.
\newblock In {\em Proceedings of the 4th International Workshop on Parallel and
  Symbolic Computation}, pages 169--170. ACM, 2010.

\bibitem{ng2004feature}
A.~Y. Ng.
\newblock Feature selection, l 1 vs. l 2 regularization, and rotational
  invariance.
\newblock In {\em Proceedings of the twenty-first international conference on
  Machine learning}, page~78. ACM, 2004.

\bibitem{lbp}
T.~Ojala, M.~Pietikainen, and D.~Harwood.
\newblock Performance evaluation of texture measures with classification based
  on kullback discrimination of distributions.
\newblock In {\em Pattern Recognition, 1994. Vol. 1-Conference A: Computer
  Vision \&amp; Image Processing., Proceedings of the 12th IAPR International
  Conference on}, volume~1, pages 582--585. IEEE, 1994.

\bibitem{puschel2005spiral}
M.~Puschel, J.~M. Moura, J.~R. Johnson, D.~Padua, M.~M. Veloso, B.~W. Singer,
  J.~Xiong, F.~Franchetti, A.~Gacic, Y.~Voronenko, et~al.
\newblock Spiral: Code generation for dsp transforms.
\newblock {\em Proceedings of the IEEE}, 93(2):232--275, 2005.

\bibitem{ren2014face}
S.~Ren, X.~Cao, Y.~Wei, and J.~Sun.
\newblock Face alignment at 3000 fps via regressing local binary features.
\newblock In {\em Proceedings of the IEEE Conference on Computer Vision and
  Pattern Recognition}, pages 1685--1692, 2014.

\bibitem{300w}
C.~Sagonas, G.~Tzimiropoulos, S.~Zafeiriou, and M.~Pantic.
\newblock A semi-automatic methodology for facial landmark annotation.
\newblock In {\em Proceedings of the IEEE Conference on Computer Vision and
  Pattern Recognition Workshops}, pages 896--903, 2013.

\bibitem{saragih2006iterative}
J.~Saragih and R.~Goecke.
\newblock Iterative error bound minimisation for aam alignment.
\newblock In {\em Pattern Recognition, 2006. ICPR 2006. 18th International
  Conference on}, volume~2, pages 1196--1195. IEEE, 2006.

\bibitem{tzimiropoulos2015project}
G.~Tzimiropoulos.
\newblock Project-out cascaded regression with an application to face
  alignment.
\newblock In {\em 2015 IEEE Conference on Computer Vision and Pattern
  Recognition (CVPR)}, pages 3659--3667. IEEE, 2015.

\bibitem{valmadre2014learning}
J.~Valmadre, S.~Sridharan, and S.~Lucey.
\newblock Learning detectors quickly with stationary statistics.
\newblock In {\em Asian Conference on Computer Vision}, pages 99--114. Springer
  International Publishing, 2014.

\bibitem{vlfeat}
A.~Vedaldi and B.~Fulkerson.
\newblock Vlfeat: An open and portable library of computer vision algorithms.
\newblock In {\em Proceedings of the 18th ACM international conference on
  Multimedia}, pages 1469--1472. ACM, 2010.

\bibitem{xiong2015global}
X.~Xiong and F.~De~la Torre.
\newblock Global supervised descent method.
\newblock In {\em Proceedings of the IEEE Conference on Computer Vision and
  Pattern Recognition}, pages 2664--2673, 2015.

\bibitem{SDM}
X.~Xiong and F.~Torre.
\newblock Supervised descent method and its applications to face alignment.
\newblock In {\em Proceedings of the IEEE conference on computer vision and
  pattern recognition}, pages 532--539, 2013.

\bibitem{300vw}
S.~Zafeiriou, G.~Tzimiropoulos, and M.~Pantic.
\newblock The 300 videos in the wild (300-vw) facial landmark tracking
  in-the-wild challenge.
\newblock In {\em ICCV Workshop}, 2015.

\end{thebibliography}

\end{document}